\pdfoutput=1

\documentclass[11pt]{article}

\usepackage[preprint]{acl}

\usepackage{times}
\usepackage{latexsym}

\usepackage[T1]{fontenc}

\usepackage[utf8]{inputenc}

\usepackage{microtype}

\usepackage{inconsolata}

\usepackage{graphicx}
\usepackage{enumitem}

\usepackage{booktabs}       
\usepackage{amsfonts}       
\usepackage{nicefrac}       
\usepackage{microtype}      
\usepackage{amsmath,amssymb}
\usepackage{algorithm}
\usepackage{algpseudocode}
\usepackage{array} 
\usepackage{graphicx} 
\usepackage{multirow} 
\usepackage{colortbl} 
\usepackage{bbding}
\usepackage{wrapfig}
\usepackage{subfigure}
\usepackage{caption}
\usepackage{enumitem}
\usepackage{titlesec}
\usepackage{makecell}
\title{\texttt{LLMSR@XLLM25}: An Empirical Study of LLM for Structural Reasoning}

\author{
  Xinye Li\textsuperscript{\dag} \quad
  Mingqi Wan\textsuperscript{\dag} \quad
  Dianbo Sui\thanks{Dianbo Sui is the corresponding author.} \\ 
  Harbin Institute of Technology \\
  \texttt{\{lixinye,2022211876\}@stu.hit.edu.cn}, \texttt{suidianbo@hit.edu.cn}
}

\begin{document}
\maketitle

\begingroup
\renewcommand\thefootnote{\dag}
\footnotetext{Equal contribution.}
\endgroup
\begin{abstract}
We present Team \textit{asdfo123}'s submission to the \texttt{LLMSR@XLLM25} shared task, which evaluates large language models on producing fine-grained, controllable, and interpretable reasoning processes.  Systems must extract all problem conditions, decompose a chain of thought into statement–evidence pairs, and verify the logical validity of each pair.  
Leveraging only the off-the-shelf \textbf{Meta-Llama-3-8B-Instruct}, we craft a concise few-shots, multi-turn prompt that first enumerates all conditions and then guides the model to label, cite, and adjudicate every reasoning step.  A lightweight post-processor based on regular expressions normalises spans and enforces the official JSON schema.  
Without fine-tuning, external retrieval, or ensembling, our method ranks \textbf{5th} overall, achieving macro-F\textsubscript{1} scores on par with substantially more complex and resource-consuming pipelines.  We conclude by analysing the strengths and limitations of our approach and outlining directions for future research in structural reasoning with LLMs. Our code is available at \url{https://github.com/asdfo123/LLMSR-asdfo123}.

\end{abstract}

\section{Introduction}

Large language models (LLMs) have recently shown impressive performance on complex reasoning tasks, spurred in part by \textit{Chain–of–Thought} (CoT) prompting, which asks the model to externalise intermediate steps before giving an answer \citep{wei2022cot}.  
Subsequent variants—such as zero-shot CoT \citep{kojima2022zeroshot}, self-consistency decoding \citep{wang2023selfconsistency}, tree-of-thought search \citep{yao2023treeofthought}, and automatically generated demonstrations \citep{zhang2022autocot}—further boost accuracy, yet these free-form rationales remain difficult to evaluate and prone to hallucinations~\citep{akbar-etal-2024-hallumeasure}.

The \texttt{LLMSR@XLLM25} shared task tackles this limitation by framing reasoning as a constrained CoT process: systems must (i) extract every explicit problem condition, (ii) segment a rationale into aligned \textbf{statement–evidence} pairs, and (iii) judge whether each evidence span \textbf{logically entails} its statement.  
Such fine-grained structure ``improves the transparency and reliability of the process'' (task description) and enables detailed diagnosis of model behaviour.  
Moreover, the step-level labels provide dense supervision for Process Reward Modeling (PRM), which optimises \emph{how} a solution is reached rather than merely \emph{what} answer is produced \citep{uesato2022stepsupervision,lightman2023letsverifystepstep}.  

Structured parsing of reasoning brings three concrete benefits.  
First, it enhances \textbf{debuggability}: developers can pinpoint the exact step where a hallucination or logical slip occurs.  
Second, it supplies explicit training signals for PRM, shown to yield more coherent and truthful solutions on mathematical benchmarks \citep{lightman2023letsverifystepstep}.  
Third, it promotes \textbf{trustworthy AI}: users can audit or amend individual steps, a requirement for safety-critical deployments and formal logic tasks such as EntailmentBank proofs \citep{dalvi2021entailmentbank} or LogicBench diagnostics \citep{parmar2024logicbench}.  

In this report we present Team \textit{asdfo123}'s lightweight submission, which relies solely on the untuned \textbf{Meta-Llama-3-8B-Instruct} \citep{meta2024llama3}.  
A compact few-shot, multi-turn prompt guides the model through all three subtasks, while a minimal post-processor enforces the official JSON schema.  
Despite its simplicity, our approach ranks \textbf{5th} overall, demonstrating that careful prompt design and constrained reasoning can rival far more elaborate pipelines.

\section{Related Work}
\subsection{Chain-of-Thought Prompting} 
Chain-of-Thought (CoT) prompting has emerged as a powerful method to enhance multi-steasoning in large language models (LLMs). Initial studies showed significant improvements by simply adding "Let's think step by step" to zero-shot prompts \citep{kojima2022zeroshot}. Self-consistency further boosts robustness by generating multiple reasoning chains and selecting the most consistent response \citep{wang2023selfconsistency}. Least-to-Most prompting addresses complex problems by decomposing them into simpler subproblems, achieving near-perfect accuracy on challenging tasks \citep{zhou2023least}.

However, CoT prompting can produce logically flawed reasoning steps, reaching correct answers through invalid logic \citep{zelikman2022star,golovneva2023roscoesuitemetricsscoring}. The Tree of Thoughts framework mitigates this by organizing reasoning into a search tree, allowing systematic backtracking and evaluation of alternative reasoning paths \citep{yao2023treeofthought}. Incorporating knowledge-graph-based verification also improves reliability \citep{he2025give,jiang2023reasoninglm}.

Recent benchmarks focus on evaluating CoT quality beyond answer accuracy, using validity and redundancy metrics to assess reasoning step-by-step \citep{xia2025evaluating,chen2025finereason}. These approaches emphasize the need for tighter integration between reasoning generation and verification.

\subsection{Parsing}
Turning natural language into structured representations is a prerequisite for dependable reasoning.  
ProgPrompt steers LLMs to emit code‐like blocks of comments, actions, and assertions for situated robot planning \citep{singh2022progprompt}.  
Self-Ask improves interpretability by decomposing a complex query into solvable sub-questions and then composing their answers \citep{press2023selfask}.  
Coupling LLMs with Answer Set Programming lets a logic engine verify every inferred rule, boosting robustness \citep{yang2023coupling}.  
RaLU aligns CoT spans with formal logic units and checks them via external solvers \citep{li2025reasoninglogic}.

For discourse‐level parsing, Rhetorical Structure Theory (RST) models text coherence via nucleus–satellite relations \citep{MANNTHOMPSON+1988+243+281}.  
Early algorithms split texts into Elementary Discourse Units and attached rhetorical relations—sometimes without explicit markers \citep{10.5555/928529}. Enhanced RST (eRST) extends this to graphs with non‐projective, concurrent relations and both implicit and explicit signals, offering more flexible, explainable structures~\citep{zeldes2024erstsignaledgraphtheory}.

\begin{figure}[!t] 
    \centering 
    \includegraphics[width=\columnwidth]{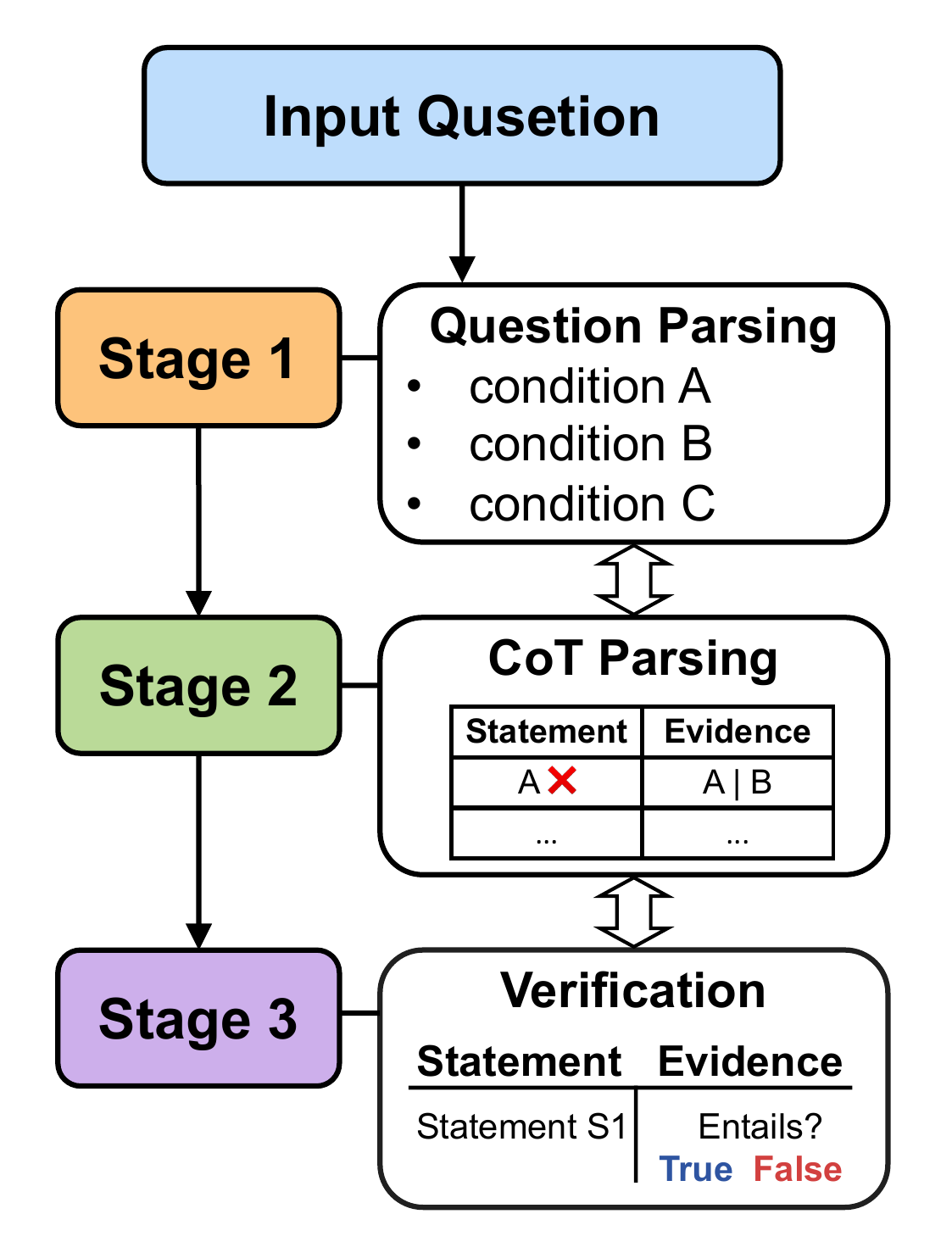}
    \caption{Illustration of the three-stage LLM-SR Task. \emph{(In our implementation, Verification is executed within the CP stage.)}}
    \label{fig:overview}
\end{figure}

\subsection{Process Reward Model}
Previous studies have demonstrated that process supervision maintains reasoning consistency better than outcome supervision, and conceptualized Process Reward Models (PRMs) to reduce logical errors~\citep{uesato2022stepsupervision,lightman2023letsverifystepstep}. To mitigate the cost of manual annotations, recent approaches automatically retrieve similar solution steps to generate fine-grained, step-level labels—facilitating both verification and PPO-based reinforcement learning without human supervision~\citep{wang2024mathshepherdverifyreinforcellms}.

Building on the foundational PRM framework, several works have further advanced process reward modeling. Tree-based preference learning constructs reasoning trees via best-first search and trains verifiers using paired step-level preferences~\citep{he2024advancingprocessverificationlarge}. More recently, CFPRM~\citep{hu2025coarsetofineprocessrewardmodeling} introduces a coarse-to-fine strategy that first merges adjacent steps into coarse-grained windows and then refines them into fine-grained units. This hierarchical method mitigates redundancy in LLM-generated reasoning while enabling training across multiple levels of granularity.

\section{Methodology}
\label{sec:method}
\subsection{Pipeline Overview}
Our system follows the three–stage workflow mandated by the LLM–SR task (Figure~\ref{fig:overview}):
\begin{enumerate}[leftmargin=*,itemsep=2pt]
    \item \textbf{Question Parsing (QP).}  
    The model enumerates every explicit condition of the problem as an ordered list.

    \item \textbf{CoT Parsing \& Verification (CP).}  
    Given the question, its Chain--of--Thought (CoT) rationale, and the QP output, the model simultaneously  
    (i) segments the rationale into \textbf{statement--evidence pairs} and  
    (ii) judges whether each evidence span \textbf{logically entails} its statement.
\end{enumerate}

All stages run on the untuned original \textbf{Meta–Llama–3–8B–Instruct}.  
Instead of parameter fine–tuning we rely on \emph{few-shot in-context learning} (ICL) with a multi-turn dialogue template (§\ref{ssec:prompt}).  
A deterministic post-processor (§\ref{ssec:post}) validates and cleans the raw generations, after which we completes the full public test set in under ten minutes.

\subsection{Prompt Engineering}
\label{ssec:prompt}

\paragraph{Few-shot demonstrations.}
We hand-pick two QP and three CP exemplars that jointly cover most patterns.  
During inference these demonstrations precede the test instance verbatim.

\paragraph{Three-turn template.}
Each call is cast as a short conversation:

\begin{enumerate}[leftmargin=1.2em,itemsep=1pt]
    \item \textsc{System}: global rules, including format restrictions.
    \item \textsc{User}: the problem text plus the explicit request (QP or CP).
    \item \textsc{Assistant}: the model’s structured JSON answer.
\end{enumerate}

Because CP depends on the extracted conditions, we invoke the model twice per instance:  
first for QP, and then for a single CP call which jointly performs CoT parsing and the verification step, with the QP list appended to the user prompt.

\subsection{Robust JSON Output}
\label{ssec:json}

Llama-3 occasionally produces ill-formed JSON—extra quotes, missing commas, or unclosed braces—which crashes the official scorer.  
By enclosing every demonstration answer in a fenced \verb|```json … ```| block and explicitly instructing the model to output valid JSON only, we cut the unparsable rate on the dev set from 16\,\% to just 2\,\%.  
The few residual errors are corrected or flagged by our post-processor.

\subsection{Post-processing}
\label{ssec:post}

A lightweight Python script performs:

\begin{enumerate}[leftmargin=*,itemsep=2pt]
    \item \textbf{Schema check}: every object must contain \texttt{statement}, \texttt{evidence}, and boolean \texttt{verification}.
    \item \textbf{Normalisation}: trim bullets, stray whitespace, smart quotes, trailing punctuation; merge duplicate conditions.
    \item \textbf{Alignment}: if \#statements $\neq$ \#evidence, align by order; otherwise flag (none observed on dev/test).
\end{enumerate}

\subsection{Efficiency Rationale}
The task rewards both answer correctness and reasoning quality.  
We show that careful prompt design plus minimal hygiene techniques already yields a top-5 macro-F\textsubscript{1} without external retrieval or fine-tuning, providing a strong, reproducible baseline for future work on PRM.

\section{Experiments}
\label{sec:exp}
We conduct all experiments on the official \texttt{LLMSR@XLLM25} test sets\footnote{\url{https://huggingface.co/datasets/shuyi-zsy/LLMSR/tree/main/llmsr}}.  
The shared task provides a \emph{fine-grained} Chain-of-Thought (CoT) analysis corpus derived from LogiQA \citep{10.5555/3491440.3491941}.  
It contains only 24 fully annotated training instances, each accompanied by both \textit{question-parsing} and \textit{CoT-parsing} labels.  
From the 24 training instances, we heuristically select a small subset of demonstrations that spans the major logical patterns; these serve as the few-shot exemplars in our prompts.

The evaluation follows a two-stage protocol.
First, we perform a $k$-shot ablation for \textbf{Question Parsing} (QP), varying the number of in-context demonstrations.
After selecting the best QP setting, we keep it fixed and sweep $k$ again for \textbf{CoT Parsing \& Verification} (CP) to determine its optimal demonstration budget.

\subsection{Phase 1: Selecting the Question‐Parsing Shot Count}
Table~\ref{tab:qp} shows QP results with $k\!\in\!\{1,2,3,4\}$.  
Macro‐F\textsubscript{1} peaks at \textbf{0.7526} with \textbf{2‐shot}.  
Adding a third or fourth example degrades performance, presumably because the longer prompt dilutes salient patterns and pushes relevant context tokens farther from the model’s attention window.

\begin{table}[h]
\centering
\begin{tabular}{@{}cc@{}}
\toprule
\textbf{Shots ($k$)} & \textbf{Question\_Macro\_F1} \\
\midrule
1 & 0.6707 \\
2 & \textbf{0.7526} \\
3 & 0.7281 \\
4 & 0.7061 \\
\bottomrule
\end{tabular}
\caption{Few‐shot ablation for Question Parsing.}
\label{tab:qp}
\end{table}

Given its clear advantage, we \emph{fix $k{=}2$ for all subsequent QP calls}.  
The extracted condition list is then passed as additional context to the CP stage.

\subsection{Phase 2: Tuning CoT Parsing \& Verification}
After fixing the QP stage at two demonstrations, we sweep the shot count for CoT Parsing.  
Table~\ref{tab:cp} shows that \textbf{3-shot} strikes the best trade-off, yielding the highest \textit{Statement\_Macro\_F1} as well as the strongest pair-level and reasoning scores.  
Adding a fourth example brings only marginal gains and in some cases degrades performance, presumably because the longer prompt pushes relevant tokens farther from the model’s attention window.

\begin{table}[h]

\centering

\setlength{\tabcolsep}{6pt}
\begin{tabular}{@{}cccc@{}}
\toprule
\textbf{CP Shots} & \textbf{Stmt}$_{\text{F1}}$ & \textbf{Stmt+Ev}$_{\text{F1}}$ & \textbf{Reasoning}$_{\text{F1}}$ \\
\midrule
1 & 0.3066 & 0.0726 & 0.0391 \\
2 & 0.1816 & 0.0860 & 0.0250 \\
3 & \textbf{0.3304} & \textbf{0.1385} & \textbf{0.0782} \\
4 & 0.2978 & 0.0976 & 0.0518 \\
\bottomrule
\end{tabular}
\caption{CoT Parsing \& Verification with \emph{2‐shot QP} fixed.  
``Stmt'' = Statement\_Macro\_F1,  
``Stmt+Ev'' = Statement\_Evidence\_Macro\_F1.}
\label{tab:cp}
   
\end{table}

\subsection{Final Configuration}
The combination of \textbf{2‐shot QP} and \textbf{3‐shot CP} constitutes our submission.  
This hybrid setup achieves the highest overall macro‐F\textsubscript{1} on the public leaderboard while preserving the system’s lightweight.
The results highlight two insights:  
(1) QP and CP favour different demonstration budgets, and  
(2) carefully tuning each stage separately beats a single fixed prompt size.

We report our final experimental results in Table~\ref{tab:phase_results}, which include the Test A and Test B phase scores on the official \texttt{LLMSR@XLLM25} test sets.

\begin{table}[h]
\centering
\resizebox{\linewidth}{!}{%
\begin{tabular}{@{}ccccc@{}}
\toprule
\textbf{Phase} & \textbf{Question$_{\text{F1}}$} & \textbf{Stmt$_{\text{F1}}$} & \textbf{Stmt+Ev$_{\text{F1}}$} & \textbf{Reasoning$_{\text{F1}}$} \\
\midrule
Test A & 75.26 & 33.04 & 13.85 & 7.82 \\
Test B & 75.33 & 47.26 & 20.17 & 11.64 \\
\bottomrule
\end{tabular}%
}
\caption{Macro-F1 scores on four evaluation criteria for Test A and Test B phases.  
``Stmt'' = Statement\_Macro\_F1,  
``Stmt+Ev'' = Statement\_Evidence\_Macro\_F1.}
\label{tab:phase_results}
\end{table}

\section{Discussion}
\label{sec:discussion}

\subsection{Key Insights from the Shared Task}
The \texttt{LLMSR@XLLM25} shared task offers a concrete sandbox for \textbf{controllable} and \textbf{transparent} reasoning.  
By forcing systems to expose every condition, align each statement with explicit evidence, and render a step-level entailment verdict, the task goes well beyond conventional answer‐only evaluation.  
Our experiments confirm three central insights:

\begin{enumerate}[leftmargin=*,itemsep=2pt]
    \item \textbf{Structural reasoning is promising yet non-trivial.}  
    Even an untuned 8B model can reliably parse conditions (\S\ref{sec:exp}, Phase~1), but struggles to decompose and verify chains of thought.
    \item \textbf{Larger does not (yet) mean satisfactory.}  
    Informal leaderboard comparisons indicate that more elaborate, resource-heavy pipelines still fall short. The bottleneck is not extraction but \textit{logical adjudication}.
\end{enumerate}

\subsection{Limitations of Llama-3-8B}
\textbf{Meta-Llama-3-8B} scores well on QP but falters on logic: it hallucinates evidence, merely paraphrases conditions, and mishandles negation, dragging down Statement–Evidence and Reasoning F\textsubscript{1}.  
These errors persist despite prompt tuning and JSON guards, implying the bottleneck lies in the model’s logic rather than the interface.

\subsection{Future Work}
\textbf{Stronger verifiers.}  
Verification may need a more capable judge (e.g., GPT-4o, Claude 3) detached from the generator.

\noindent
\textbf{Lightweight entailment modules.}  
Training a small, dedicated critic on synthetic entailment pairs—à la CoT-Critic—could boost step-level faithfulness.

\noindent
\textbf{Process Reward Models (PRMs).}  
The extracted structures are ideal supervisory signals for PRMs. Iteratively refining the generator with PRM feedback may tighten the link between evidence and statements, increasing coherence without brute-force scaling.

\subsection{Takeaway}
The shared task shows that structured reasoning is a feasible yet unsolved frontier for LLMs.  
Our minimal system serves as a proof of concept; progress now hinges on developing (i) stronger or specialised verifiers and (ii) learning paradigms that reward \emph{how} a conclusion is reached, not merely \emph{what} it is.  
We believe these directions will be pivotal for deploying LLMs in settings where transparency and trustworthiness are non-negotiable.

\section{Conclusion}
We showed that a carefully crafted, few-shot prompting pipeline—backed by lightweight post-processing—can tackle the \texttt{LLMSR@XLLM25} shared task without fine-tuning or external tools, ranking 5$^{\text{th}}$ overall.  
While Meta-Llama-3-8B handles condition extraction well, its verification accuracy remains limited, underscoring the need for stronger or specialised reasoners and process-level training signals.  
Future work should pair stronger base models with dedicated entailment critics and reward models that explicitly value step-by-step correctness.

\bibliography{custom}

\begin{thebibliography}{29}
\providecommand{\natexlab}[1]{#1}

\bibitem[{Akbar et~al.(2024)Akbar, Hossain, Wood, Chin, Salinas, Alvarez, and
  Cornejo}]{akbar-etal-2024-hallumeasure}
Shayan~Ali Akbar, Md~Mosharaf Hossain, Tess Wood, Si-Chi Chin, Erica~M Salinas,
  Victor Alvarez, and Erwin Cornejo. 2024.
\newblock \href {https://doi.org/10.18653/v1/2024.emnlp-main.837}
  {{H}allu{M}easure: Fine-grained hallucination measurement using
  chain-of-thought reasoning}.
\newblock In \emph{Proceedings of the 2024 Conference on Empirical Methods in
  Natural Language Processing}, pages 15020--15037, Miami, Florida, USA.
  Association for Computational Linguistics.

\bibitem[{Chen et~al.(2025)Chen, Xu, Zhang, Chan, Liu, Bing, Zhao, Luu, and
  Rong}]{chen2025finereason}
Guizhen Chen, Weiwen Xu, Hao Zhang, Hou~Pong Chan, Chaoqun Liu, Lidong Bing,
  Deli Zhao, Anh~Tuan Luu, and Yu~Rong. 2025.
\newblock Finereason: Evaluating and improving llms’ deliberate reasoning
  through reflective puzzle solving.
\newblock arXiv preprint arXiv:2502.20238.

\bibitem[{Dalvi et~al.(2021)Dalvi, Jansen, Tafjord, Xie, Smith, Pipatanangkura,
  and Clark}]{dalvi2021entailmentbank}
Bhavana Dalvi, Peter Jansen, Oyvind Tafjord, Zhengnan Xie, Hannah Smith,
  Leighanna Pipatanangkura, and Peter Clark. 2021.
\newblock \href {https://doi.org/10.18653/v1/2021.emnlp-main.585} {Explaining
  answers with entailment trees}.
\newblock In \emph{Proceedings of the 2021 Conference on Empirical Methods in
  Natural Language Processing}, pages 7358--7370, Online and Punta Cana,
  Dominican Republic. Association for Computational Linguistics.

\bibitem[{Golovneva et~al.(2023)Golovneva, Chen, Poff, Corredor, Zettlemoyer,
  Fazel-Zarandi, and Celikyilmaz}]{golovneva2023roscoesuitemetricsscoring}
Olga Golovneva, Moya Chen, Spencer Poff, Martin Corredor, Luke Zettlemoyer,
  Maryam Fazel-Zarandi, and Asli Celikyilmaz. 2023.
\newblock \href {https://arxiv.org/abs/2212.07919} {Roscoe: A suite of metrics
  for scoring step-by-step reasoning}.
\newblock \emph{Preprint}, arXiv:2212.07919.

\bibitem[{He et~al.(2025)He, Ma, Fan, Roth, Wang, and Ribeiro}]{he2025give}
Jiashu He, Mingyu~Derek Ma, Jinxuan Fan, Dan Roth, Wei Wang, and Alejandro
  Ribeiro. 2025.
\newblock Give: Structured reasoning of large language models with knowledge
  graph inspired veracity extrapolation.
\newblock arXiv preprint arXiv:2410.08475.

\bibitem[{He et~al.(2024)He, Shen, Zhang, Tan, and
  Lu}]{he2024advancingprocessverificationlarge}
Mingqian He, Yongliang Shen, Wenqi Zhang, Zeqi Tan, and Weiming Lu. 2024.
\newblock \href {https://arxiv.org/abs/2407.00390} {Advancing process
  verification for large language models via tree-based preference learning}.
\newblock \emph{Preprint}, arXiv:2407.00390.

\bibitem[{Hu et~al.(2025)Hu, Chen, Zhao, Ouyang, and
  Liu}]{hu2025coarsetofineprocessrewardmodeling}
Yulan Hu, Ge~Chen, Jinman Zhao, Sheng Ouyang, and Yong Liu. 2025.
\newblock \href {https://arxiv.org/abs/2501.13622} {Coarse-to-fine process
  reward modeling for mathematical reasoning}.
\newblock \emph{Preprint}, arXiv:2501.13622.

\bibitem[{Jiang et~al.(2023)Jiang, Zhou, Zhao, Li, and
  Wen}]{jiang2023reasoninglm}
Jinhao Jiang, Kun Zhou, Xin Zhao, Yaliang Li, and Ji-Rong Wen. 2023.
\newblock Reasoninglm: Enabling structural subgraph reasoning in pre-trained
  language models for question answering over knowledge graph.
\newblock In \emph{EMNLP}, pages 3721--3735.

\bibitem[{Kojima et~al.(2022)Kojima, Gu, Reid, Matsuo, and
  Iwasawa}]{kojima2022zeroshot}
Takeshi Kojima, Shixiang~Shane Gu, Machel Reid, Yutaka Matsuo, and Yusuke
  Iwasawa. 2022.
\newblock Large language models are zero-shot reasoners.
\newblock In \emph{Proceedings of the 36th International Conference on Neural
  Information Processing Systems}, NIPS '22, Red Hook, NY, USA. Curran
  Associates Inc.

\bibitem[{Li et~al.(2025)Li, Xu, and Guo}]{li2025reasoninglogic}
Cheryl Li, Tianyuan Xu, and Yiwen Guo. 2025.
\newblock Reasoning-as-logic-units: Scaling test-time reasoning in large
  language models through logic unit alignment.
\newblock arXiv preprint arXiv:2502.07803.

\bibitem[{Lightman et~al.(2023)Lightman, Kosaraju, Burda, Edwards, Baker, Lee,
  Leike, Schulman, Sutskever, and Cobbe}]{lightman2023letsverifystepstep}
Hunter Lightman, Vineet Kosaraju, Yura Burda, Harri Edwards, Bowen Baker, Teddy
  Lee, Jan Leike, John Schulman, Ilya Sutskever, and Karl Cobbe. 2023.
\newblock \href {https://arxiv.org/abs/2305.20050} {Let's verify step by step}.
\newblock \emph{Preprint}, arXiv:2305.20050.

\bibitem[{Liu et~al.(2021)Liu, Cui, Liu, Huang, Wang, and
  Zhang}]{10.5555/3491440.3491941}
Jian Liu, Leyang Cui, Hanmeng Liu, Dandan Huang, Yile Wang, and Yue Zhang.
  2021.
\newblock Logiqa: a challenge dataset for machine reading comprehension with
  logical reasoning.
\newblock In \emph{Proceedings of the Twenty-Ninth International Joint
  Conference on Artificial Intelligence}, IJCAI'20.

\bibitem[{MANN and THOMPSON(1988)}]{MANNTHOMPSON+1988+243+281}
WILLIAM~C. MANN and SANDRA~A. THOMPSON. 1988.
\newblock \href {https://doi.org/doi:10.1515/text.1.1988.8.3.243} {Rhetorical
  structure theory: Toward a functional theory of text organization}.
\newblock \emph{Text - Interdisciplinary Journal for the Study of Discourse},
  8(3):243--281.

\bibitem[{Marcu(1998)}]{10.5555/928529}
Daniel~C. Marcu. 1998.
\newblock \emph{The rhetorical parsing, summarization, and generation of
  natural language texts}.
\newblock Ph.D. thesis, CAN.
\newblock AAINQ35238.

\bibitem[{{Meta AI}(2024)}]{meta2024llama3}
{Meta AI}. 2024.
\newblock \href {https://huggingface.co/meta-llama/Meta-Llama-3-8B-Instruct}
  {Meta-llama-3-8b-instruct model card}.

\bibitem[{Parmar et~al.(2024)Parmar, Patel, Varshney, Nakamura, Luo, Mashetty,
  Mitra, and Baral}]{parmar2024logicbench}
Mihir Parmar, Nisarg Patel, Neeraj Varshney, Mutsumi Nakamura, Man Luo, Santosh
  Mashetty, Arindam Mitra, and Chitta Baral. 2024.
\newblock \href {https://doi.org/10.18653/v1/2024.acl-long.739}
  {{L}ogic{B}ench: Towards systematic evaluation of logical reasoning ability
  of large language models}.
\newblock In \emph{Proceedings of the 62nd Annual Meeting of the Association
  for Computational Linguistics (Volume 1: Long Papers)}, pages 13679--13707,
  Bangkok, Thailand. Association for Computational Linguistics.

\bibitem[{Press et~al.(2023)Press, Zhang, Min, Schmidt, Smith, and
  Lewis}]{press2023selfask}
Ofir Press, Muru Zhang, Sewon Min, Ludwig Schmidt, Noah Smith, and Mike Lewis.
  2023.
\newblock Measuring and narrowing the compositionality gap in language models.
\newblock In \emph{Findings of EMNLP}, pages 5687--5711.

\bibitem[{Singh et~al.(2022)Singh, Blukis, Mousavian, Goyal, Xu, Tremblay, Fox,
  Thomason, and Garg}]{singh2022progprompt}
Ishika Singh, Valts Blukis, Arsalan Mousavian, Ankit Goyal, Danfei Xu, Jonathan
  Tremblay, Dieter Fox, Jesse Thomason, and Animesh Garg. 2022.
\newblock Progprompt: Generating situated robot task plans using large language
  models.
\newblock arXiv preprint arXiv:2209.11302.

\bibitem[{Uesato et~al.(2022)Uesato, Kushman, Kumar, Song, Siegel, Wang,
  Creswell, Irving, and Higgins}]{uesato2022stepsupervision}
Jonathan Uesato, Nate Kushman, Ramana Kumar, Francis Song, Noah Siegel, Lisa
  Wang, Antonia Creswell, Geoffrey Irving, and Irina Higgins. 2022.
\newblock \href {https://arxiv.org/abs/2211.14275} {Solving math word problems
  with process- and outcome-based feedback}.
\newblock \emph{Preprint}, arXiv:2211.14275.

\bibitem[{Wang et~al.(2024)Wang, Li, Shao, Xu, Dai, Li, Chen, Wu, and
  Sui}]{wang2024mathshepherdverifyreinforcellms}
Peiyi Wang, Lei Li, Zhihong Shao, R.~X. Xu, Damai Dai, Yifei Li, Deli Chen,
  Y.~Wu, and Zhifang Sui. 2024.
\newblock \href {https://arxiv.org/abs/2312.08935} {Math-shepherd: Verify and
  reinforce llms step-by-step without human annotations}.
\newblock \emph{Preprint}, arXiv:2312.08935.

\bibitem[{Wang et~al.(2023)Wang, Wei, Schuurmans, Le, Chi, Narang, Chowdhery,
  and Zhou}]{wang2023selfconsistency}
Xuezhi Wang, Jason Wei, Dale Schuurmans, Quoc Le, Ed~Chi, Sharan Narang,
  Aakanksha Chowdhery, and Denny Zhou. 2023.
\newblock \href {https://arxiv.org/abs/2203.11171} {Self-consistency improves
  chain of thought reasoning in language models}.
\newblock \emph{Preprint}, arXiv:2203.11171.

\bibitem[{Wei et~al.(2023)Wei, Wang, Schuurmans, Bosma, Ichter, Xia, Chi, Le,
  and Zhou}]{wei2022cot}
Jason Wei, Xuezhi Wang, Dale Schuurmans, Maarten Bosma, Brian Ichter, Fei Xia,
  Ed~Chi, Quoc Le, and Denny Zhou. 2023.
\newblock \href {https://arxiv.org/abs/2201.11903} {Chain-of-thought prompting
  elicits reasoning in large language models}.
\newblock \emph{Preprint}, arXiv:2201.11903.

\bibitem[{Xia et~al.(2025)Xia, Li, Liu, Wu, and Liu}]{xia2025evaluating}
Shijie Xia, Xuefeng Li, Yixin Liu, Tongshuang Wu, and Pengfei Liu. 2025.
\newblock Evaluating mathematical reasoning beyond accuracy.
\newblock arXiv preprint arXiv:2404.05692.

\bibitem[{Yang et~al.(2023)Yang, Ishay, and Lee}]{yang2023coupling}
Zhun Yang, Adam Ishay, and Joohyung Lee. 2023.
\newblock Coupling large language models with logic programming for robust and
  general reasoning from text.
\newblock arXiv preprint arXiv:2307.07696.

\bibitem[{Yao et~al.(2023)Yao, Yu, Zhao, Shafran, Griffiths, Cao, and
  Narasimhan}]{yao2023treeofthought}
Shunyu Yao, Dian Yu, Jeffrey Zhao, Izhak Shafran, Thomas~L. Griffiths, Yuan
  Cao, and Karthik Narasimhan. 2023.
\newblock \href {https://arxiv.org/abs/2305.10601} {Tree of thoughts:
  Deliberate problem solving with large language models}.
\newblock \emph{Preprint}, arXiv:2305.10601.

\bibitem[{Zeldes et~al.(2024)Zeldes, Aoyama, Liu, Peng, Das, and
  Gessler}]{zeldes2024erstsignaledgraphtheory}
Amir Zeldes, Tatsuya Aoyama, Yang~Janet Liu, Siyao Peng, Debopam Das, and Luke
  Gessler. 2024.
\newblock \href {https://arxiv.org/abs/2403.13560} {erst: A signaled graph
  theory of discourse relations and organization}.
\newblock \emph{Preprint}, arXiv:2403.13560.

\bibitem[{Zelikman et~al.(2022)Zelikman, Wu, Mu, and
  Goodman}]{zelikman2022star}
Eric Zelikman, Yuhuai Wu, Jesse Mu, and Noah Goodman. 2022.
\newblock Star: Bootstrapping reasoning with self-consistency.
\newblock In \emph{Advances in Neural Information Processing Systems},
  volume~35, pages 15476--15488.

\bibitem[{Zhang et~al.(2022)Zhang, Zhang, Li, and Smola}]{zhang2022autocot}
Zhuosheng Zhang, Aston Zhang, Mu~Li, and Alex Smola. 2022.
\newblock \href {https://arxiv.org/abs/2210.03493} {Automatic chain of thought
  prompting in large language models}.
\newblock \emph{Preprint}, arXiv:2210.03493.

\bibitem[{Zhou et~al.(2023)Zhou, Schärli, Hou, Wei, Scales, Wang, Schuurmans,
  Cui, Bousquet, Le, and Chi}]{zhou2023least}
Denny Zhou, Nathanael Schärli, Le~Hou, Jason Wei, Nathan Scales, Xuezhi Wang,
  Dale Schuurmans, Claire Cui, Olivier Bousquet, Quoc Le, and Ed~Chi. 2023.
\newblock \href {https://arxiv.org/abs/2205.10625} {Least-to-most prompting
  enables complex reasoning in large language models}.
\newblock \emph{Preprint}, arXiv:2205.10625.

\end{thebibliography}

\appendix



\end{document}